\documentclass{article}

\usepackage[colorlinks,citecolor=gray,linkcolor=black]{hyperref} 

\usepackage[preprint]{neurips_2023}

\usepackage[utf8]{inputenc} 
\usepackage[T1]{fontenc}    
\usepackage{hyperref}       
\usepackage{url}            
\usepackage{booktabs}       
\usepackage{amsfonts}       
\usepackage{nicefrac}       
\usepackage{microtype}      
\usepackage{xcolor}         
\usepackage{xspace}

\usepackage{amsmath}
\usepackage{amssymb}
\usepackage{mathtools}
\usepackage{amsthm}
\usepackage{multirow}
\usepackage{subcaption}

\usepackage{float}

\newcommand{\model}{ModuleFormer\xspace}
\newcommand{\lm}{MoLM\xspace}

\title{ModuleFormer:\\ Modularity Emerges from Mixture-of-Experts}

\author{%
  Yikang Shen\thanks{Correspondence: \texttt{yikang.shen@ibm.com}}\\
  MIT-IBM Watson AI Lab\\
  \And
  Zheyu Zhang \\
  Tsinghua University \\
  \And
  Tianyou Cao \\
  Tsinghua University \\
  \AND
  Shawn Tan \\
  Mila/University of Montreal \\
  \And
  Zhenfang Chen \\
  MIT-IBM Watson AI Lab\\
  \And
  Chuang Gan \\
  MIT-IBM Watson AI Lab\\
}

\begin{document}

\maketitle

\begin{abstract}
Large Language Models (LLMs) have achieved remarkable results. 
But existing models are expensive to train and deploy, and it is also difficult to expand their knowledge beyond pre-training data without forgetting previous knowledge. 
This paper proposes a new neural network architecture, \textit{\model}, that leverages modularity to improve the efficiency and flexibility of large language models. 
\model is based on the Sparse Mixture of Experts (SMoE).
Unlike previous SMoE-based modular language model~\citep{gururangan2021demix}, which requires domain-labeled data to learn domain-specific experts, \model can induce modularity from uncurated data with its new load balancing and load concentration losses. 
\model is a modular architecture that includes two different types of modules, new stick-breaking attention heads, and feedforward experts. 
Different modules are sparsely activated conditions on the input token during training and inference.
In our experiment, we found that the modular architecture enables three important abilities for large pre-trained language models:
1) Efficiency, since \model only activates a subset of its modules for each input token, thus it could achieve the same performance as dense LLMs with more than \textbf{two times} throughput;
2) Extendability, \model is more immune to catastrophic forgetting than dense LLMs and can be easily extended with new modules to learn new knowledge that is not included in the training data;
3) Specialisation, finetuning \model could specialize a subset of modules to the finetuning task, and the task-unrelated modules could be easily pruned for a lightweight deployment. 
The inference code and model weights are here: \url{https://github.com/IBM/ModuleFormer}.

\end{abstract}

\section{Introduction}
While modern Large Language Models (LLMs) have achieved remarkable results and even surpassed human performance on some tasks, it remains inefficient and inflexible.  
Most LLMs (e.g. Llama, \citealt{touvron2023llama}; Pythia, \citealt{biderman2023pythia}; GPT-3, \citealt{brown2020language}) use all of their parameters during inference and training. 
We refer to these as dense models.
However, previous work has shown that, a large portion of parameters in a neural model can be pruned while still maintaining similar performance \citep{frankle2018lottery,frantar2023sparsegpt,lecun1989optimal}.
Furthermore, LLMs are `frozen in time' once trained, but many practical use cases require the LLMs to have up-to-date knowledge~\citep{lazaridou2022internet}.

Furthermore, fine-tuning the entire model for domain adaptation or continual learning is becoming costly and compute-prohibitive as model sizes grow, making it infeasible for users with smaller computational budgets.
Updating all parameters also makes the model vulnerable to catastrophic forgetting \citep{mccloskey1989catastrophic, aghajanyan2021muppet}.
To this end, lightweight adaptation methods like LoRA that update only a small subset of the original parameters are becoming popular \citep{hu2021lora,samarakoon2016low}.
However, our experiments show that such methods can still suffer catastrophic forgetting and are less capable of adapting to a substantially different domain, like a new language.

Modularity could be a good solution for LLMs to address the aforementioned issues. 
A modular model could have several benefits:
1) the model could activate a subset of module conditions on the input or task, thus using less computation than densely activating the entire model;
2) given a domain or task, a subset of domain/task-related modules could be assembled to form a new lightweight model;
3) the model could be easily extended with new modules for domain adaption or continual learning;
4) the model could be more immune to catastrophic forgetting because only the input-related modules are updated during the model fine-tuning.

The notion of neural network modules is not new. 
\citet{andreas2016neural} proposes Neural Module Networks (NMN) for visual question-answering tasks.
NMN has limited practicality because it requires intensive domain knowledge to assign a predefined functionality for each neural module and combines these modules with respect to the internal structure of the input question. 
\textsc{DEMix} \citep{gururangan2021demix} and Mod-Squad \citep{chen2022mod} propose to leverage Sparse Mixture of Experts [SMoEs, \citealt{shazeer2017outrageously}] to combine modularity with the Transformer architecture.
While experts in SMoE are similar to the modules in NMN, the drawback of SMoE is the lack of established methods to manipulate experts, including selecting experts for a specific task and adding new experts for domain adaptation. 
To solve this problem, \textsc{DEMix}  and Mod-Squad use curated training data to learn the functionality of each expert.
However, curated data is always hard to obtain and scale up.

\begin{figure}[t]
    \centering
    \begin{subfigure}{0.285\textwidth}
        \centering
        \includegraphics[width=\textwidth]{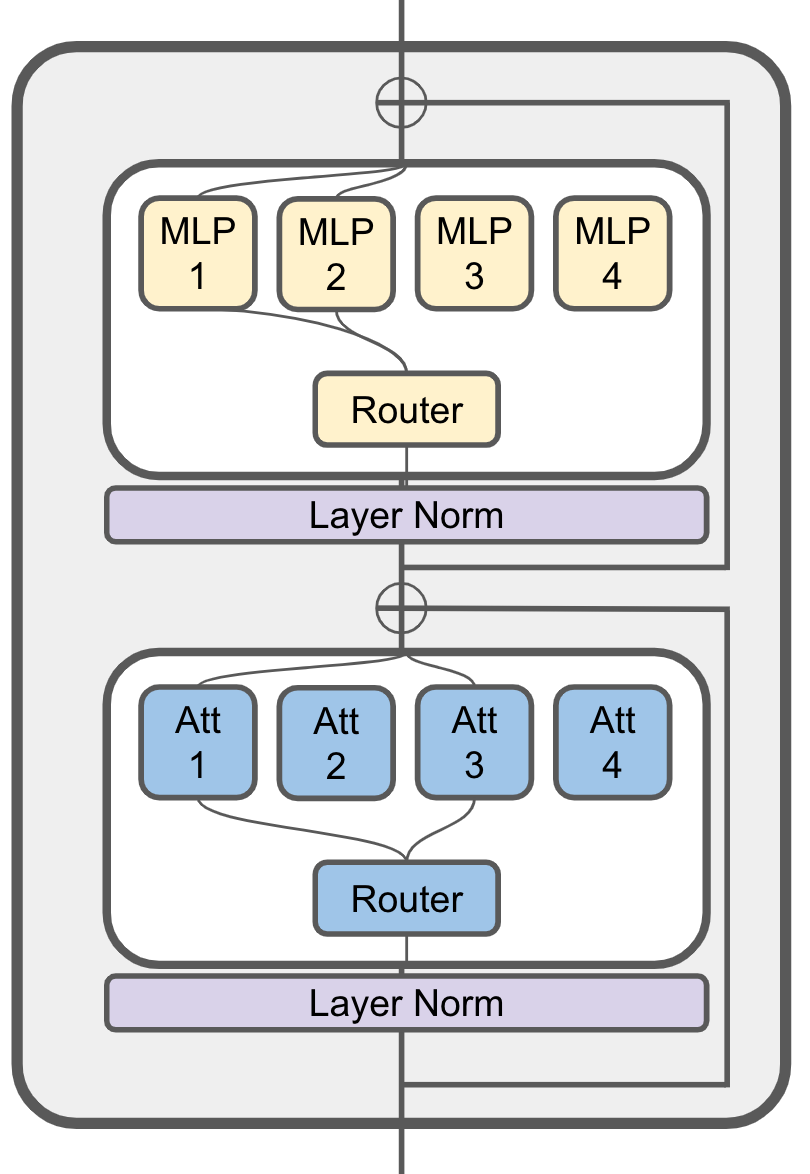}
        \caption{Sparse activation of modules}
        \label{fig:architecture}
    \end{subfigure}
    \quad
    \begin{subfigure}{0.335\textwidth}
        \centering
        \includegraphics[width=\textwidth]{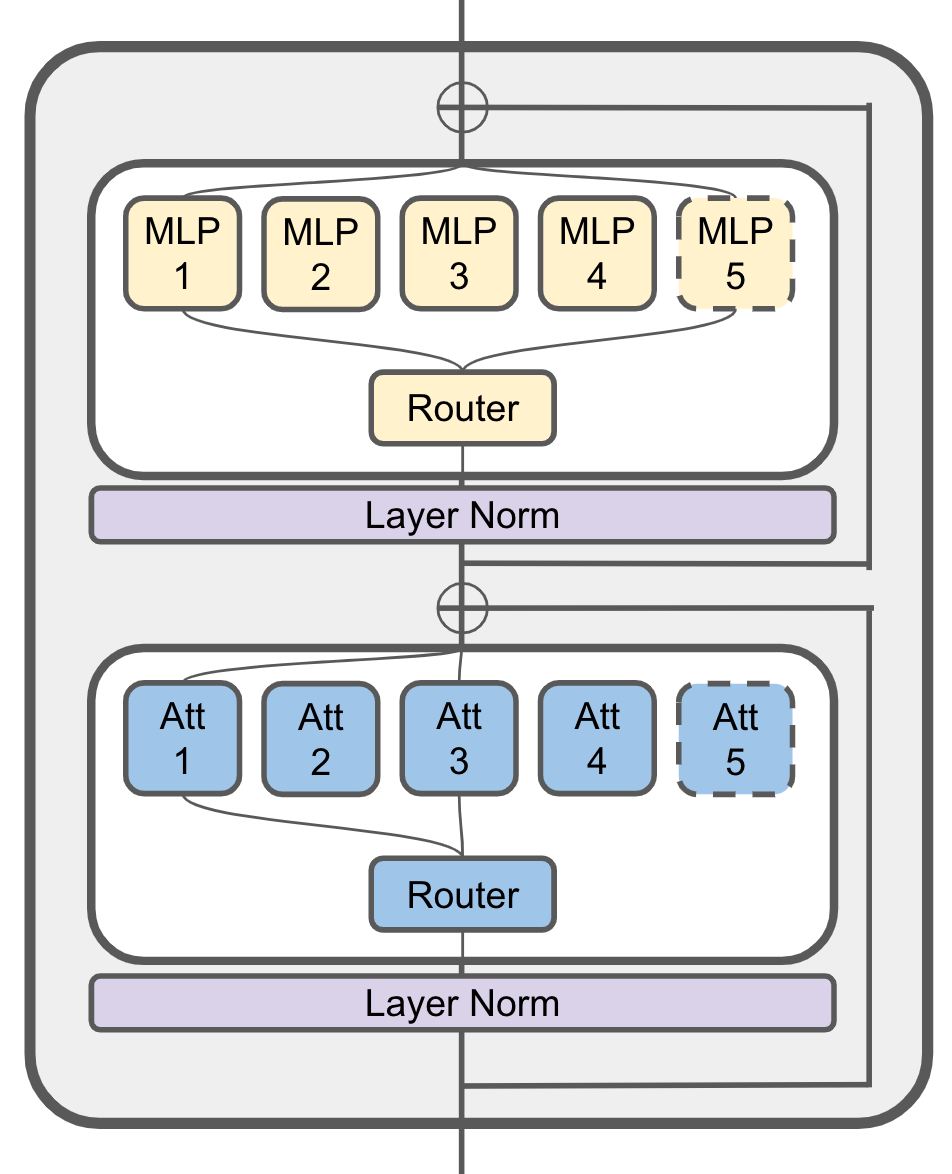}
        \caption{Adding new modules}
        \label{fig:new_modules}
    \end{subfigure}
    \quad
    \begin{subfigure}{0.29\textwidth}
        \centering
        \includegraphics[width=\textwidth]{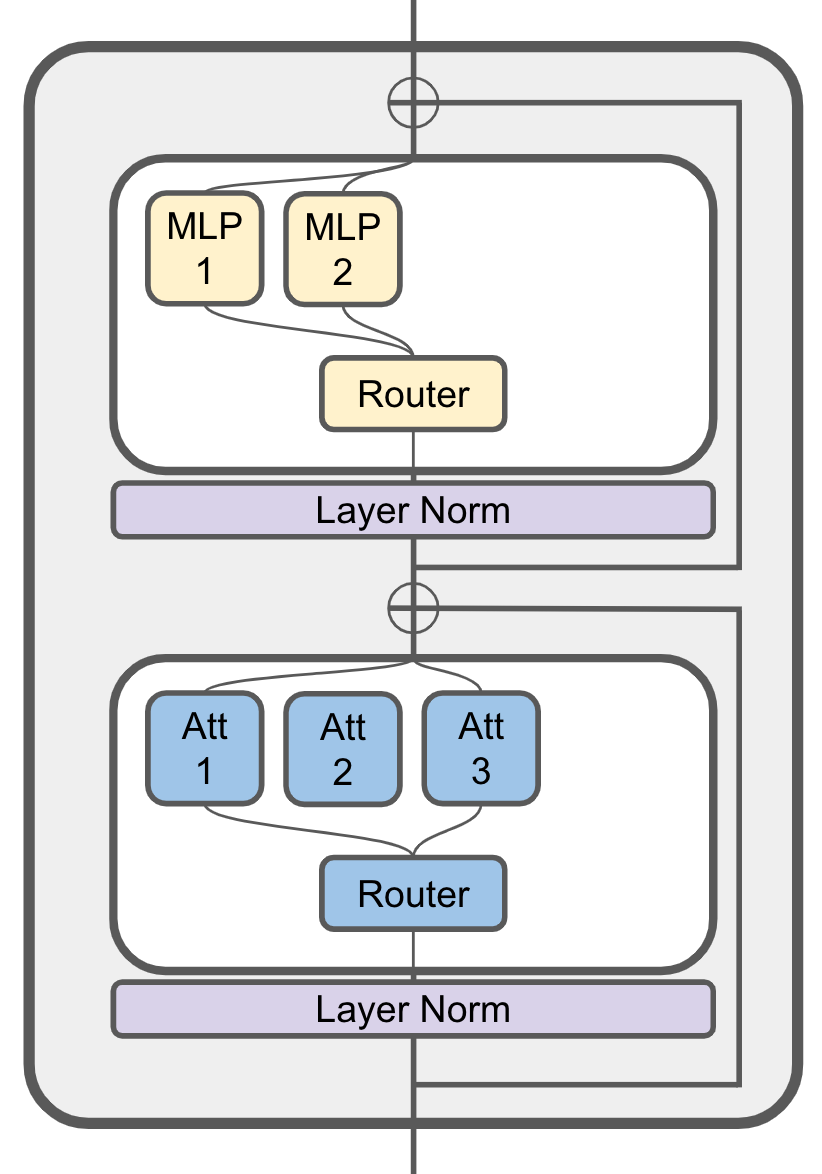}
        \caption{Pruning modules}
        \label{fig:pruning}
    \end{subfigure}
    \caption{The architecture of \model. 
    The sparse activation schema enables high computation efficiency.
    Adding new modules is simply inserting randomly initialized ones into each layer and then training the new experts and the router on a new dataset.
    The number and size of new modules can be customized to accommodate different scenarios.
    Pruning modules involves counting the activation frequency of each module and setting a threshold to remove the least used modules.
    The percentage of prune could also be customized to achieve a trade of between performance and model size.}
    \vspace{-.25in}
\end{figure}

This paper argues that modularity can emerge from language model pretraining with uncurated data.
We propose a new modular architecture, \textsc{\model} (Fig.~\ref{fig:architecture}), and the associated methods for module manipulation. 
\model comprises FFD modules and a novel stick-breaking attention mechanism, 
and a new mutual information loss that balances the load of different modules.
Furthermore, we demonstrate inserting new modules (Fig.~\ref{fig:new_modules}), and expert pruning (Fig.~\ref{fig:pruning}) can be done in \model.
To enable pruning, we introduce a new load concentration loss to select and specialize a subset of modules on a given task.
Our experiment result shows the promising abilities of \model: 
1) It achieves the same performance as dense LLMs with lower latency (50\%) and a smaller memory footprint; thus, it could process more than 2 times the number of tokens per second;
2) It is less susceptible to catastrophic forgetting after finetuning the entire model on a new domain, and it could also be easily extended with new modules to learn a new language;
3) It can be finetuned on a downstream task to specialize a subset of modules on the task and the unused modules can be pruned without sacrificing performance.

\section{Related Work}
\textbf{Pretrained Language Models and finetuning.}
Pretraining is crucial for attaining good results on NLP tasks. 
In previous transfer-learning regimes, the \textit{de facto} standard was to finetune a pretrained model on a downstream task, updating all its parameters  \citep{devlin2018bert}. 
With the increasing scale of language models \citep{radford2019language,brown2020language,chowdhery2022palm}, this procedure has become increasingly difficult for researchers without similar computing resources to the research groups that trained them.
As a result, prompting and prompt engineering became the go-to method for NLP researchers to leverage the capabilities of these LLMs to solve their task \citep{brown2020language}. 
Alternatively, methods that adapt only a subset of parameters grew in popularity \citep{houlsby2019parameter,rebuffi2017learning,lin2020exploring}.
\emph{Low-rank adaptation} (LoRA, \citealt{hu2021lora}) is a popular method being used in a wide variety of models. 
We show that LoRA, while effective for small adaptations to LLMs for specific tasks, fail to perform as well if tasked to learn a different language. 

\textbf{Sparse Mixture of Experts for Modularity}
SMoE methods for sparse conditional computation was introduced in \citet{shazeer2017outrageously}, primarily with the goal of scaling up models. 
\citet{fedus2021switch} used this property to great effect to train the Switch Transformer, sharding the different experts to different compute nodes to fully leverage distributed training regimes.
Our goal, however, is to leverage SMoEs for the purpose of \emph{modularity}.
The notion of modularity in neural models is not new.
Previous works have proposed several ways to introduce modularity to neural network models. 
\citet{andreas2016neural} introduces the Neural Module Network (NMN) for visual question-answering tasks. 
An NMN is composed of several predefined neural network modules.
Each module has a clear definition of its functionality.
These modules are dynamically assembled for different instances of a reasoning task.
But NMN requires highly curated data with expert labels, and the number of predefined modules is task-specific, thus limiting its application to other tasks.

\textsc{DEMix} \citep{gururangan2021demix} combines the SMoE with modular schema. 
It proposes to train different feedforward experts for different domains, with each feedforward expert functioning as a module with specific domain knowledge.
However, \textsc{DEMix} is pretrained on curated data with domain labels, whereas the state-of-the-art LLMs are trained on large-scale domain-agnostic data.
Furthermore, when the input domain is unknown, the inference cost of Demix linearly increases with the number of modules.

\textbf{Length-extrapolation in Attention}
Position embeddings introduced in \citet{vaswani2017attention} used sinusoidal embeddings to represent positions in the sentence.
This should allow the key-query interactions to extrapolate to positions unseen during training, but this is unfortunately not the case in practice \citep{press2021train}.
Efforts have been made to improve the extrapolation capabilities of position embeddings \citep{su2021roformer,sun2022length}.

In \model, we encode positional information without positional embeddings.
Our method can be considered as a simplification of Geometric Attention \citep{csordas2021neural}. 
This formulation, also known as \emph{stick-breaking} after this formulation of the Dirichlet process, has been used in different settings in deep learning \citep{dehghani2018universal,banino2021pondernet,graves2016adaptive,tan2016towards}.
While ALiBi~\citep{press2021train} also biases toward recent time steps, ALiBi can inhibit attending on terms further away, as the positional bias overwhelms the content attention scores, whereas stick-breaking attention does not have the same issue.

\textbf{Continual Learning} Prior work has demonstrated the advantages of continual learning \citep{alsentzer-etal-2019-publicly, chakrabarty-etal-2019-imho, lee2020biobert, beltagy2019scibert} .
When data that was previously used in training is available, joint training~\citep{caruana1997multitask} is known to be the most straightforward method with the best performance~\citep{li2017learning}.
For lifelong learning (i.e. previously trained data is not available), the newly trained network tends to forget knowledge learned in the previously trained tasks.
This is known as \emph{catastrophic forgetting}~\citep{mccloskey1989catastrophic}.
To overcome this phenomenon, \cite{kirkpatrick2017overcoming} proposed a regularization method.
\cite{munkhdalai2017meta, beaulieu2020learning} learns to perform lifelong learning through meta-learning.
These approaches are orthogonal to methods like \textsc{DEMix}-dapt~\citep{gururangan2021demix} and \model, and can be combined with these methods.
However, \textsc{DEMix}-dapt  belongs to the \textit{duplicated and fine-tuned} category in the taxonomy introduced by \cite{li2017learning}, where a model is duplicated several times for each task.
This causes great inconvenience when performing inference on domain-agnostic data.

\textbf{Neural Network Pruning}
Neural Network Pruning eliminates redundant parameters and structure to accelerate inference in neural networks.
Previous work has focused more on structured model pruning~\citep{li2016pruning, he2017channel}. These pruned models often benefit from additional finetuning to achieve improved performance. However, other recent work show that compressed models can be trained from scratch to achieve comparable performance, eliminating the need for the reliance on fine-tuning~\citep{liu2018rethinking, wang2020pruning}.
One way pruning is simplified in SMoE-based models is to prune off entire experts, without significantly changing model architecture.
However, previous SMoE-based large language models~\citep{shazeer2017outrageously,lepikhin2020gshard,fedus2021switch} have been trained with load balancing losses which forces the model to use every expert.
As a result, pruning experts is more likely to lead to sacrifices in performance.
We propose the load concentration loss that biases the model towards using fewer modules during adaptation or finetuning to alleviate this problem.


\section{Model Architecture}
\subsection{Preliminary: Mixture of Experts}
\newcommand{\x}{\mathbf{x}}
\newcommand{\key}{\mathbf{k}}
\newcommand{\val}{\mathbf{v}}
\newcommand{\query}{\mathbf{q}}
\newcommand{\out}{\mathbf{o}}
\newcommand{\W}{\mathbf{W}}
\newcommand{\y}{\mathbf{y}}
\newcommand{\A}{\mathbf{A}}
\newcommand{\B}{\mathbf{B}}

A Mixture of Experts (MoE) layer comprises $N$ modules $f_1,\hdots,f_N$ and a router $g(m \mid \x)$.
Given an input $\x$ to the MoE layer, the router predicts a probability distribution over the $N$ modules.
Of these, we select the top $k$ experts.
When $k < N$ 
, we are using a Sparse Mixture of Experts (SMoE, \citealt{shazeer2017outrageously}). 
In this paper, we use an MLP to model the router
\begin{align}
    g(m \mid \x) &= \text{Top}k \left( \frac{e^{h(\x)_i}}{\sum e^{h(\x)_j}} \right) \label{eq: moe}, \\
    h(\x) &= \A\ \mathrm{ReLU}(\B \x) \label{eq:gating},
\end{align}
where $h(\x)$ is modeled by an MLP, $\A$ is the expert embedding matrix of shape $(N,D_{\text{rtr}})$, $\B$ is the input projection matrix of shape $(D_\text{rtr},D_\text{emb})$, $\text{Top}k$ is the operator that sets all gates to zero except the top $k$ gates.
The final output of the SMoE is then given by $y = \sum_{m=1}^N g(m \mid \x) \cdot f_m(\x)$. 
When $g(m \mid \x) = 0$,  $f_m(\x)$ will not need to be evaluated, thus reducing computation cost during training and inference.

Compared to a standard transformer decoder block, a \model block replaces the Feed-forward layer (FFD) with a SMoE layer and the self-attention layer with a Mixture of Attention heads (MoA) layer inspired by \cite{zhang2022mixture}.
Each SMoE layer can be described by a 3-tuple, $(N_\mathrm{ffd}, k, D_\mathrm{ffd})$, with $N_\mathrm{ffd}$ as the number of FFD modules, $k$ is the parameter in the top-$k$ operation, and $D_\mathrm{ffd}$ is the dimension of the hidden layer inside each module.
Like the MoE layer, an MoA layer has $N_\mathrm{att}$ attention modules and activates the top-$k$ modules for each input.
Unlike \citet{zhang2022mixture}, we introduce a new stick-breaking self-attention module.

\subsection{Stick-breaking Self-Attention head}


The stick-breaking self-attention is designed for the Transformer decoder to model the attention of each token $\x_t$ to previous tokens $\x_{<t}$.
It uses the stick-breaking process view of the Dirichlet process to model the attention distribution instead of the softmax in a standard attention layer.
The motivation to pay attention to the latest matching tokens. 
It can also be considered a simplification of the geometric attention proposed in \cite{csordas2021neural}.

Given an input vector sequence of $t$ time steps $\x_1, \x_2, ..., \x_t$, each input is projected to a sequence of key vectors $\key_1, \key_2, ..., \key_t$ and a sequence of value vectors $\val_1, \val_2, ..., \val_t$.
To compute the attention of time step $t$, the input $\x_t$ is projected to a query vector $\query_t = \W_q \x_t$, where $\W_q$ is the query projection matrix. 
For all previous steps and the current step $i \leq t$, we compute the probability that the key at time step $i$ matches the query at time step $t$:
\begin{equation}
    \beta_{i,t} = \mathrm{sigmoid}(\key_i^{\mathsf{T}}  \query_t).
\end{equation}
To get the attention weights of the most recent matching key, we use the stick-breaking process:
\begin{equation}
    p_{i,t} = \beta_{i,t} \prod_{i < j \leq t} \left( 1 - \beta_{j,t} \right). \label{eq:stick-breaking}
\end{equation} 
Note that $\sum_i p_{i,t} \leq 1$, and sums to 1 given a sufficiently long context (See Appendix \ref{sec:sbproof}).
Further, as in \cite{csordas2021neural}, this can be efficiently computed with a combination of cumulative sums in log-space.
Based on the attention distribution $p_{i,t}$, we can compute the attention output as:
\begin{equation}
    \out_t = \W_o^{\mathsf{T}} \sum_{i \leq t} p_{i,t} \val_i,
\end{equation}
where $\W_o$ is the output projection matrix.

Since Equation~\ref{eq:stick-breaking} selects the latest match token, the stick-breaking attention implicitly encodes position information, negating the need for explicit modeling of position with sinusoidal embeddings or relative position biases, and drastically simplifies length-extrapolation of self-attention. 
After pre-training on a fixed input length, the model could theoretically process any input length.
Like \cite{dai2019transformer},  we also concatenate the key and value from the previous batch to the current batch during pre-training, extending the effective context length from sequence length $T$ to $N_L T$, where $N_L$ is the number of ModuleFormer Blocks.

As in \cite{zhang2022mixture}, each stick-breaking self-attention head is composed of four $\mathbb{R}^{D_{emb} \times D_{att}}$ matrix $\W_q, \W_k, \W_v, \W_o$, where $D_{att}$ is the attention head dimension, $\W_k$ and $\W_v$ are shared by different heads, $\W_q$ and $\W_o$ are different across heads.
In the case of $D_{att}$ equal to $D_{emb}$, we can set $\W_k$ and $\W_v$ to the identity matrix to eliminate the shared parameters between heads.

\section{Module Manipulation}
\subsection{Load Balancing during Pretraining}
To avoid the SMOEs repeatedly using the same module and wasting the extra capacity in the other modules, it requires various load balancing losses to regulate the training of the router \citep{shazeer2017outrageously,fedus2021switch}.
Mod-Squad \citep{chen2022mod} proposes to maximize the Mutual Information between modules and tasks to balance the load of different experts and build alignment between tasks and experts.

Unlike \cite{chen2022mod}, we want to maximize the Mutual Information (MI) between tokens and modules:
\begin{equation}
\mathcal{L}_\text{MI} = \sum_{\x} \sum_m p(m,\x) \log \left(\frac{p(m, \x)}{p(m)p(\x)} \right).
\end{equation}
We assume for simplicity that $\x$ is uniform over the set of $\x$ in the batch, $\mathcal{X}$, and therefore $p(\x) = \frac{1}{\mathcal{X}}$.
After removing all constant components, we can simplify the MI loss to be the difference between the entropy of $p(m)$ and the conditional entropy of $p(m \mid \x)$:
\begin{equation}
   \mathcal{L}_\text{MI} = \underbrace{\sum_{m=1}^N  p(m) \log p(m)}_{-H(m)} - \frac{1}{|\mathcal{X}|} \sum_{\x \in \mathcal{X}} \underbrace{ \sum_{m=1}^N  g(m\mid \x) \log g(m\mid \x)}_{H(m \mid \x)},
\end{equation}
where $p(m)=\sum_{\x} g(m| \x)p(\x)$, $p(\x)$ is the probability of each token inside the batch, $H(m)$ is the entropy of modules' marginal distribution, $H(m \mid \x)$ is the entropy the modules' probability condition on input $\x$, $|\mathcal{X}|$ is the number of tokens.
For a minibatch of size $B$ with length $T$, the number of tokens is $|\mathcal{X}| = BT$, and the token probability is $p(x) = 1/|\mathcal{X}|$.
Intuitively, the MI loss maximizes the entropy of modules' marginal distribution and minimizes the entropy of the conditional distribution of modules given an input $\x$. 
It balances the load of each expert across the entire batch (maximize $H(m)$), but also encourages each input $\x$ to concentrate their gating probability to fewer modules (minimize $H(m \mid \x)$).

\subsection{Load Concentration during Finetuning} 
\label{sec:load_concentration}
While we want to maximize the use of each expert during pretraining, 
we want to hone in on the modules used for specific downstream tasks.
This way, we can remove unused modules and reduce the number of parameters for the finetuned model.
To concentrate the load on fewer experts, we minimize the marginal entropy instead,
\begin{equation}
    \mathcal{L}_\text{entropy} = H(m) = - \sum_{m=1}^N  p(m) \log p(m),
\end{equation}
encouraging the model to use fewer experts.
After fine-tuning, we can count the module frequency $f_m$ used on the training or validation sets. 
The $f_m$ represents the importance of module $m$ for this task.  
We can easily prune the model by removing experts with $f_m$ less than a certain threshold.
Appendix~\ref{app: ablation: pruning} demonstrates the benefits of load concentration when finetuning.

\subsection{Inserting new Modules for Continual Learning}

For modularized models, inserting new modules is a straightforward and parameter-efficient method to learn new knowledge without fine-tuning the entire model.
When inserting $N_{new}$ randomly initialized modules to each layer, we also extend the module embedding layer $\A$ in the router~(Eq.~\ref{eq:gating}) with a new matrix $\A'$ of shape $(N_{new}, D_\text{rtr})$. 
Hence the new routing function can be written as:
\begin{equation}
h'(\x) = \begin{bmatrix}
    \A \\
    \A'
\end{bmatrix}\ \mathrm{ReLU}(\B \x).
\end{equation}

Since the parameters in previous modules are frozen during fine-tuning, continual learning with new modules could largely avoid the catastrophic forgetting problem. 
However, catastrophic forgetting could still happen to the router. 
This occurs when new modules are trained on a new domain, and the router erroneously routes input from the old domain to the new expert. 

To avoid this issue, we make the router partially trainable and apply regularization. 
In detail, we freeze  $\A$ and $\B$, leaving only $\A'$ trainable. 
Furthermore, we introduce a technique called 'routing regularization', which restricts the norm of $\A'$:
\begin{equation}
\mathcal L_{\mathrm{rout\_reg}} = \big|\big|\A'\big|\big|^2.
\end{equation}
This method effectively limits the usage of new experts, as a larger magnitude of $\A'$ corresponds to a higher probability of the new experts being prioritized.

Unlike classical regularization approaches for continual learning, such as decay or L2 Loss, which have been pointed out to be defective~\citep{lesort2019regularization}, our routing regularization doesn't restrict the capacity of experts but only restricts the tendency of usage of new experts.
Appendix~\ref{app: ablation: continual} demonstrates the benefits of the partially trainable router and routing regularization.


%

\section{Experiments}
\subsection{Language Modeling}

\begin{table}[b]
    \centering
    \small
    \caption{Language Model Hyperparameters}
    \begin{tabular}{c|cccccccccc}
        \toprule
        \multirow{2}{*}{Model} & \multirow{2}{*}{$D_{\text{emb}}$} & \multirow{2}{*}{$N_\mathrm{layer}$} & \multirow{2}{*}{$N_\mathrm{att}$} & \multirow{2}{*}{$D_\mathrm{att}$} & \multirow{2}{*}{$N_\mathrm{ffd}$} & \multirow{2}{*}{$D_\mathrm{ffd}$} & \multirow{2}{*}{Top$k$} & total & active \\
         & & & & & & & & params & params\\
        \midrule
        \lm-4B-K2 & 1024 & 24 & 16 & 1024 & 32 & 2048 & 2 & 4B & 350M \\
        \lm-4B-K4 & 1024 & 24 & 16 & 1024 & 32 & 2048 & 4 & 4B & 700M \\
        \lm-8B-K2 & 1024 & 48 & 16 & 1024 & 32 & 2048 & 2 & 8B & 700M \\
        \bottomrule
    \end{tabular}
    \label{tab:hyperparameters}
    \vspace{-0.2in}
\end{table}

\paragraph{Pretraining}
We pretrained three different versions of the \model Language Model (\lm).
The hyperparameters of different models are charted in Table~\ref{tab:hyperparameters}.
These models are pretrained on the Pile corpus \citep{gao2020pile}, whose training corpus contains roughly 300 billion tokens.
We tokenize the corpus with the Codegen tokenizer~\citep{nijkamp2022codegen}.
All models are trained with the AdamW optimizer \citep{loshchilov2017decoupled} with a maximum learning rate of 3e-4.
We use a cosine learning rate schedule with a warmup of 3 billion tokens.
The initial and final learning rate is equal to 10\% of the maximal learning rate, with a weight decay of 0.01 and gradient clipping of 1.0.
Each training batch is about 3 million tokens, and the sequence length is 512 tokens.
Like the Transformer-XL \citep{dai2019transformer}, we also concatenate the attention key and value of the previous batch to the current batch, resulting in a context length of 1024.
We optimize the training efficiency with Pytorch Fully Sharded Data Parallel (FSDP).
It takes 48 A100 GPUs with 80GB of RAM and 6 days to train the \lm-4B-K2 model.
More pre-training details can be found in Appendix~\ref{app:pretrain}.

\paragraph{Evaluation Settings}
In keeping with previous work, we report results on a total of 8 tasks, including 0-shot, few-shot, language modeling, and code generation tasks. We compare \lm with other open-source language models of similar computations, size, and training data, including Pythia~\citep{biderman2023pythia} and GPT-Neo~\citep{gao2020pile}.
All three families of models are trained on the Pile dataset, which allows us to focus on the comparison of the model architecture.

We use the Language Model Evaluation Harness\footnote{\url{https://github.com/EleutherAI/lm-evaluation-harness}} to evaluate language models on zero-shot, few-shot, and language modeling tasks.
For the zero-shot and few-shot tasks, the objective is to select the most appropriate completion among a set of given options based on a provided context. 
We select the completion with the highest likelihood given the provided context.
For language modeling, we test on the Wikitext dataset.
The objective is to minimize the perplexity of the next token prediction.
For code generation, we evaluate models on the HumanEval dataset~\citep{chen2021evaluating}.
HumanEval contains 164 hand-written Python programming problems. 
The model needs to complete a function given the task description prompt such that it can pass all provided test cases. 

\paragraph{Results}
\begin{table}[t]
    \small
    \caption{Inference Speed and Zero-shot performance. 
    We measure the end-to-end latency and peak memory consumption for an input of batch size 32 and length 1024 on an A100 80GB GPU.
    We also measure the throughput of each model when the GPU memory is fit with the maximum number of 1024-token sequences. 
    }
    \label{tab:zero_shot}
    \centering
    \begin{tabular}{c|cc|c|ccccc}
        \toprule
         & Latency & Memory & Throughput & Hellaswag & PIQA & ARC-e & ARC-c & OBQA \\
         & ms & GB & tokens/sec&  \multicolumn{5}{c}{acc} \\
        \midrule
        Pythia 410M & 554 & 25 & 59594 & 33.72 & 66.70 & 51.89 & 21.42 & 18.2 \\
        \midrule
        GPT-Neo 1.3B & 991 & 23 & 32857 & 38.66 & 71.11 & 56.19 & 23.12 & 21.4 \\
        Pythia 1.4B & 918 & 42 & 35559 & 40.41 & 70.84 & 60.52 & 26.11 & 22.2 \\
        \textbf{\lm-4B-K2} & 497 & 27 & 71017 & 39.21 & 70.13 & 56.44 & 23.55 & 20.8  \\
        \midrule
        GPT-Neo 2.7B & 1737 & 35 & 18788 & 42.71 & 72.2 & 61.07 & 27.47 & 23.2  \\
        Pythia 2.8B & 2111 & 70 & 15522 & 45.34 & 73.99 & 64.35 & 29.35 & 23.8  \\
        \textbf{\lm-4B-K4} & 863 & 27 & 39931 & 42.20 & 73.01 & 60.82 & 25.94 & 22.6 \\
        \textbf{\lm-8B-K2} & 939 & 38 & 37419 & 43.33 & 72.91 & 62.46 & 27.90 & 23.8 \\
        \bottomrule
    \end{tabular}
    \vspace{-.2in}
\end{table}


\begin{table}[]
    \small
    \centering
    \caption{Few Shot, Code Generation, and Language Modeling performance.}
    \label{tab:others}
    \begin{tabular}{c|ccc|ccc|c}
        \toprule
             &  \multicolumn{3}{c|}{TriviaQA} & \multicolumn{3}{c|}{HumanEval pass@k {[}\%{]}} & Wikitext\\
             & 0-shot & 1-shot & 5-shot & k=1 & k=10 & k=100 & PPL\\
            \midrule
            Pythia 410M & 2.32 & 5.02 & 6.42 & 1.20 & 3.85 & 9.98 & 20.09 \\
            \midrule
            GPT-Neo 1.3B & 5.24 & 8.01 & 9.74 & 3.62 & 6.87 & 14.50 & 16.16 \\
            Pythia 1.4B & 5.30 & 9.87 & 12.84 & 2.19 & 7.31 & 14.33 & 14.71\\
            \textbf{\lm-4B-K2} & 5.40 & 11.12 & 13.70 & 3.04 & 6.99 & 13.79 & 15.15 \\
            \midrule
            GPT-Neo 2.7B & 4.82 & 11.23 & 13.67 & 4.89 & 9.54 & 17.90 & 13.93 \\
            Pythia 2.8B & 7.38 & 15.58 & 18.98 & 4.91 & 11.76 & 21.54 & 12.68\\ 
            \textbf{\lm-4B-K4} & 9.07 & 14.24 & 16.49 & 5.50 & 10.65 & 20.27 & 13.20 \\
            \textbf{\lm-8B-K2} & 11.47 & 16.73 & 20.75 & 5.51 & 12.58 & 20.40 & 12.97 \\
            \bottomrule
        \end{tabular}
        \vspace{-0.2in}
\end{table}



Table~\ref{tab:zero_shot} and Table~\ref{tab:others} show the performance of \lm and baseline language models on common sense reasoning, closed-book question answering, and code generation benchmarks.
Overall, \lm-4B-K2 model achieves comparable performance with dense models of around 1.3 billion parameters, \lm-4B-K4 and \lm-8B-K2 model achieve comparable performance with dense models of around 2.7 billion parameters.
Thanks to its sparse computation schema, 
\lm only uses around 25\% active parameters per token compared to its dense counterpart. 
Consequently, it reduces the latency to 50\% while having lower peak memory usage, and also increases the throughput by 2 times when the GPU memory of fully occupied.

\paragraph{Ablation Study}
We implemented the ablation study on stick-breaking attention and mutual information loss. All models are trained on a 100b tokens subset of the Pile corpus. The results are presented in Table~\ref{tab:ablation}.

The ablation results show that our stick-breaking attention outperforms the RoPE-based attention used in state-of-the-art LLMs like LLama~\citep{touvron2023llama}. 
During our training, we also noticed that the original self-attention with learnable absolute position embedding is unstable. 
It spikes during training and fails to recover. 
In contrast, RoPE-based attention and stick-breaking attention are more stable.
To study the effectiveness of Mutual Information loss, we compared it with two baselines: 1) no load balancing loss and 2) the load balancing loss used in ST-MoE~\citep{zoph2022st}, which is an extension of the load balancing loss proposed in \cite{shazeer2017outrageously}.
We found that the MI loss outperforms the model without load-balancing loss on all tasks and achieves comparable performance as ST-MoE loss. 

\begin{table}[t]
    \centering
    \begin{tabular}{l|ccccccc}
        \toprule
        \multirow{2}{*}{Model} & Hellaswag & PIQA & ARC-e & ARC-c & OBQA & Wikitext \\
        & acc & acc & acc & acc & acc & PPL \\
        \midrule
        MoLM-4B-K2 & \textbf{37.10} & \textbf{70.24} & 55.09 & 23.98 & 21.2 & \textbf{17.47} \\
        - Stick-breaking + Softmax\&RoPE & 35.66 & 67.85 & 54.17 & 22.35 & 20.4 & 19.92 \\
        - MI loss & 35.64 & 69.21 & 53.41 & 23.29 & 19.2 & 19.36 \\
        - MI loss + ST-MoE loss & 36.93 & 69.42 & \textbf{55.6} & \textbf{24.15} & \textbf{22.8} & 18.20 \\
        \bottomrule
    \end{tabular}
    \caption{Ablation study for self-attention mechanism and load balancing loss. ``- Stick-breaking + Softmax\&RoPE'' means that we replace stick-breaking attention with softmax self-attention and RoPE~\citep{su2021roformer}. ``- MI loss'' means that we remove the Mutual Information load balancing loss. ``- MI loss + ST-MoE loss'' means that we replace MI loss with the load balancing loss used in ST-MoE~\citep{zoph2022st}.}
    \label{tab:ablation}
\end{table}

\begin{figure}[]
    \centering
	\begin{subfigure}{0.43\textwidth}
		\centering
		\includegraphics[width=\textwidth]{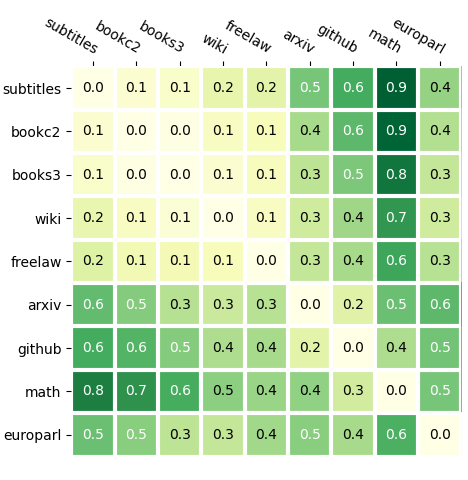}
		\caption{Mutual Information Loss}
	\end{subfigure}
	\quad
	\begin{subfigure}{0.52\textwidth}
		\centering
		\includegraphics[width=\textwidth]{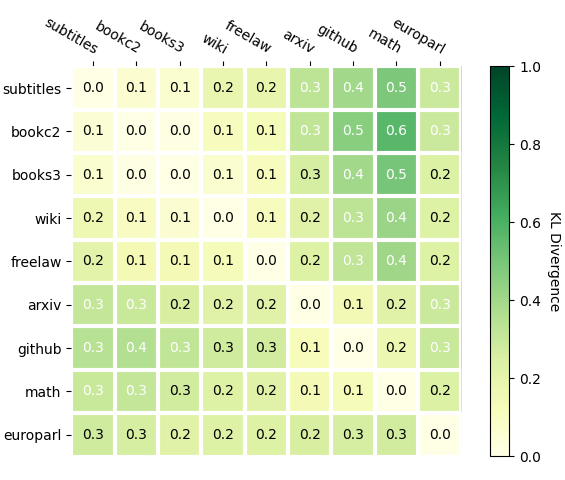}
		\caption{ST-MoE loss}
	\end{subfigure}
    \caption{KL-divergence between different domains of Pile test set. We collected expert activation frequencies for MLP experts of MoLM-4B-K2 and our ST-MoE baseline on different domains of the Pile test set. We computed the KL-divergency between domains from these expert distributions for both MoLM-4B-K2 and ST-MoE baseline. Lower KL-divergence means similar expert distribution for two domains.}
    \label{fig:kl-divergence}
\end{figure}

\paragraph{Analysis of expert distribution}
We further collected expert distribution for MLP experts of MoLM-4B-K2 and our ST-MoE baseline on different domains of the Pile test set. 
Figure~\ref{fig:kl-divergence} shows the KL-divergency of expert distribution between domains.
We noticed two interesting phenomenons:
1) Similar domains have smaller KL divergence;
2) ST-MoE has smaller KL-divergence values compared to MoLM model, which suggests more confounding of experts between domains. 
This result suggests that MI loss encourages a stronger correlation between the input category and experts, thus better at incentive modularity than the ST-MoE load balancing loss.

\subsection{Learning New Knowledge with New Modules} \label{sec: new lan}



\paragraph{Experiment Settings}


In this section, we study two experiment settings: continual joint pre-training (Section~\ref{sec: joint train}) and continual lifelong pre-training (Section~\ref{sec: fine tune}). 
The difference lies in the presence or absence of English texts. 
We continually pre-train \model and GPT-Neo for both settings on languages in CC-100 Corpus~\citep{wenzek-etal-2020-ccnet, conneau-etal-2020-unsupervised}.
To evaluate quality, we employ the mLAMA benchmark~\citep{kassner2021multilingual} with the zero-shot setting used in XGLM~\citep{lin2022few} and mGPT~\citep{shliazhko2022mgpt}.
More details can be found in Appendix~\ref{app: continual details}.

\begin{table}[!t]
  \caption{Continual Joint Pre-Training Result (accuracy$\uparrow$).}
  \label{tab: ver2}
  \centering
  \small
  \begin{tabular}{lcc|cc|cc}
    \toprule
    & \multirow{2}{*}{Model} & Trainable & \multicolumn{2}{c|}{Continual Training (de)} & \multicolumn{2}{c}{Continual
    Training (vi)} \\
    & & Params & en & de & en & vi \\
    \midrule
    \multirow{2}{*}{Freeze All} & GPT-Neo-2.7B & 0 & 78.6 & 64.5 & 78.6 & \textbf{58.9} \\
    & \textbf{\lm-4B-K4} & 0 & \textbf{80.9} & \textbf{67.8} & \textbf{80.9} & 58.4 \\
    \midrule
    \multirow{2}{*}{Full Finetune} & GPT-Neo-2.7B & 2.7B & 75.1(-3.5) & 71.0(+6.5) & 75.1(-3.5) & 66.0(+7.1)\\
    & \textbf{\lm-4B-K4} & 4B & \textbf{80.1}(-0.8) & \textbf{71.5}(+3.7) & \textbf{80.1}(-0.8) & \textbf{67.9}(+9.5) \\
    \midrule
     LoRA & GPT-Neo-2.7B & 159M & 75.8(-2.8) & 66.3(+1.8) & 75.3(-3.3) & 60.4(+1.5) \\
    Insert New Experts & \textbf{\lm-4B-K4} & 164M & \textbf{79.5}(-1.4) & \textbf{69.9}(+2.1) & \textbf{79.5}(-1.4) & \textbf{64.5}(+6.1) \\
    \bottomrule \\
  \end{tabular}
\end{table}

\begin{table}[t]
  \small
  \caption{Continual Lifelong Pre-Training Result (accuracy$\uparrow$).}
  \label{tab: zero en origin}
  \centering
  \small
  \begin{tabular}{lccc|cc}
    \toprule
    & \multirow{2}{*}{Model} & Trainable & \multirow{2}{*}{Regularization} & \multicolumn{2}{c}{Continual Training (vi)} \\
    & & Params & & en & vi \\
    \midrule
    \multirow{2}{*}{Freeze All} & GPT-Neo-2.7B & 0 & N/A & 78.6 & \textbf{58.9} \\
    & \textbf{\lm-4B-K4} & 0 & N/A & \textbf{80.9} & 58.4 \\
    \midrule
    \multirow{3}{*}{LoRA} & GPT-Neo-2.7B & 133M & N/A & 70.3(-8.3) & 59.8(+1.4) \\
     & GPT-Neo-2.7B & 24M & N/A & 69.3(-9.3) & 57.5(-1.4) \\
     & GPT-Neo-2.7B & 133M & 0.25 Weight Decay & 69.9(-8.7) & 57.8(-0.6) \\
     \midrule
    \multirow{4}{*}{Insert New Experts} & \textbf{\lm-4B-K4} & 151M & $0.00$ Rout Reg & 74.5(-6.4) & \textbf{68.0}(+9.6) \\
     & \textbf{\lm-4B-K4} & 151M & $0.25$ Rout Reg & 76.0(-4.9) & 64.8(+6.4) \\
     & \textbf{\lm-4B-K4} & 151M & $0.50$ Rout Reg & 76.3(-4.6) & 64.5(+6.1) \\
     & \textbf{\lm-4B-K4} & 151M & $1.00$ Rout Reg & \textbf{78.7}(-2.2) & 63.0(+4.6) \\
    \bottomrule 
  \end{tabular}
\end{table}


\paragraph{Continual Joint Pre-Training} \label{sec: joint train}

In this part, we perform continual pre-training on models with joint training.
Specifically, we mixed English and a new language to form a new training corpus, and kept the embedding layer trainable.
Joint Training~\citep{caruana1997multitask} is a well-known method for multitask learning that demonstrates proficiency in both old and new tasks~\citep{chen2018continual}. 
However, it often creates negative interference between different tasks~\citep{gupta2022sparsely}.


Table~\ref{tab: ver2} presents the results obtained from the continually trained models.
The table reveals the following findings:
1) Consistent with previous studies~\citep{gupta2022sparsely}, we observed that sparse models experience less interference, ultimately leading to better full finetune performance;
2) In terms of efficient tuning, our proposed \model also demonstrates consistently better results compared to the baseline. This suggests that low interference comes mainly from the sparse architecture rather than a large number of trainable parameters.





\paragraph{Continual Lifelong Pre-Training} \label{sec: fine tune}

For these experiments, the models are trained only on new language texts.
\cite{abraham2005memory} proposed the stability-plasticity dilemma that explains a difficult challenge for models: 1) models should have high plasticity to learn a new language, 2) models must possess exceptional stability, considering it would not be exposed to any English tokens through the  numerous training iterations.

Table~\ref{tab: zero en origin} shows the results of the LoRA baseline and our method with different weights of the routing regularization loss. 
Our \model exhibits a strong ability to balance and trade-off stability-plasticity with the help of the routing regularization loss. 
When we restrict the usage of new experts by increasing the loss weight, the model gains stability and loss in plasticity. 
In contrast, fine-tuning GPT-Neo with LoRA falls behind on both stability and plasticity.
For comparison, LoRA is worse on both plasticity and stability. Reducing the number of trainable parameters (lower-rank) and applying weight decay to LoRA weights won't improve neither plasticity or stability.



\subsection{Finetuning and Pruning Modules}
Previous results show that \lm has specialized experts for different domains. 
In this section, we demonstrate that \lm can be pruned to create a domain-specialized model by pruning the unused experts. 
The resulting model is significantly smaller in size but maintains similar performance.

We propose two pruning strategies: 1) pruning by activation frequency first and then fine-tuning on the target domain; 2) fine-tuning on the target domain with load concentration loss and then pruning by activation frequency.
The first strategy achieves better overall performance for different pruning ratios. 
However, the second strategy only requires one fine-tuning process to enable different pruning ratios, which enables dynamically adjusting model size during inference.

\begin{figure}[t]
    \centering
	\begin{subfigure}{0.3\textwidth}
		\centering
		\includegraphics[width=\textwidth]{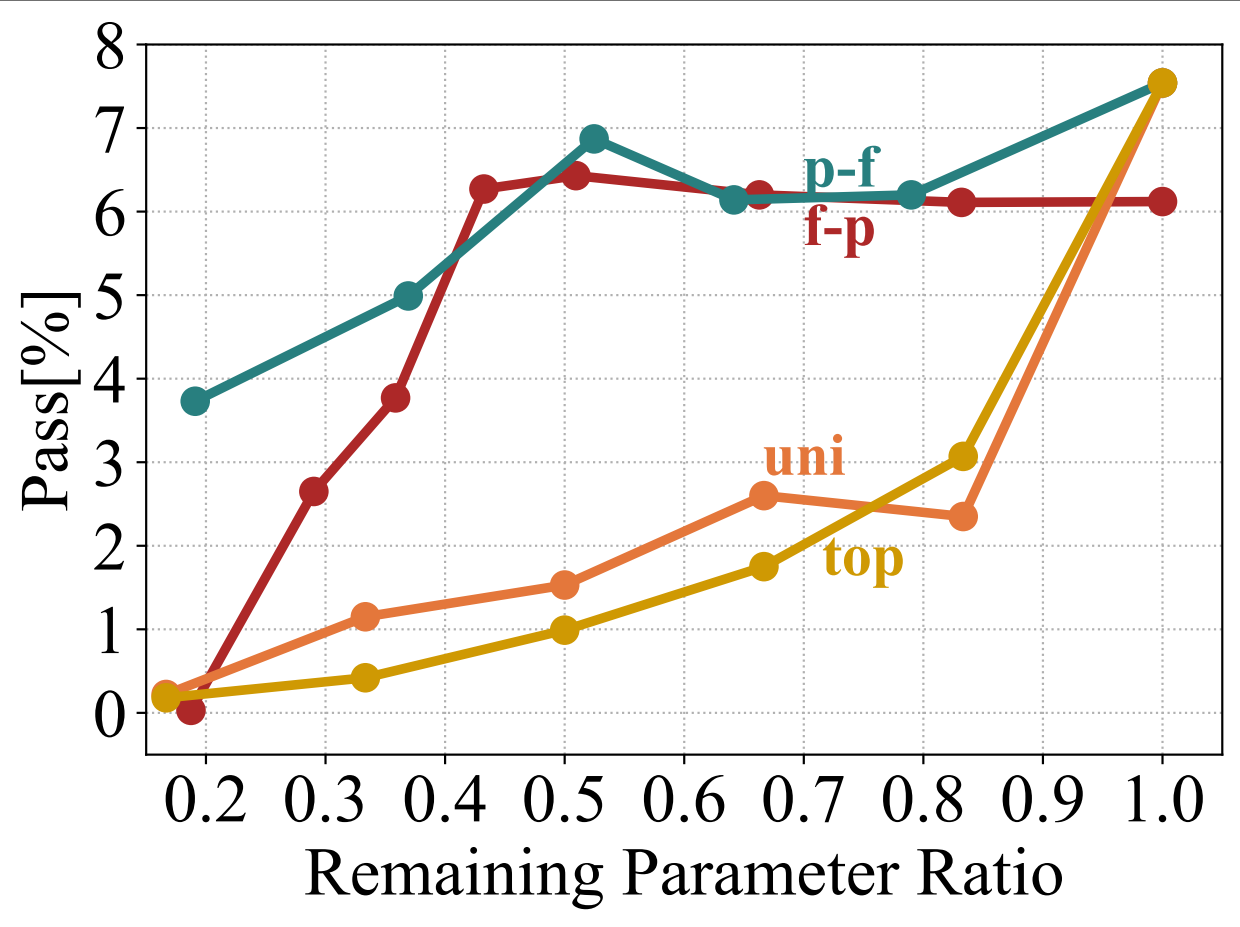}
		\caption{Pass@1}
	\end{subfigure}
	\quad
	\begin{subfigure}{0.3\textwidth}
		\centering
		\includegraphics[width=\textwidth]{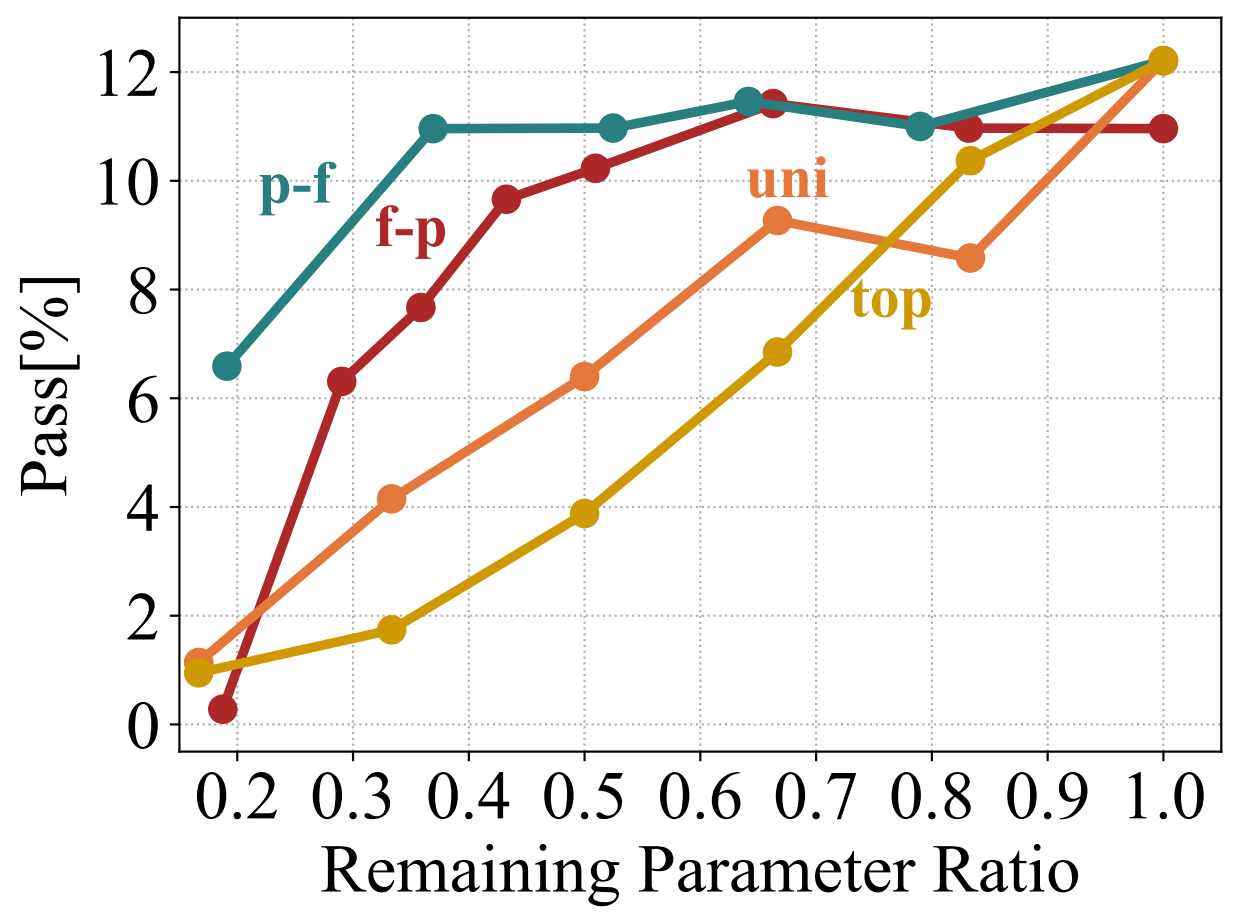}
		\caption{Pass@10}
	\end{subfigure}
	\quad
	\begin{subfigure}{0.3\textwidth}
		\centering
		\includegraphics[width=\textwidth]{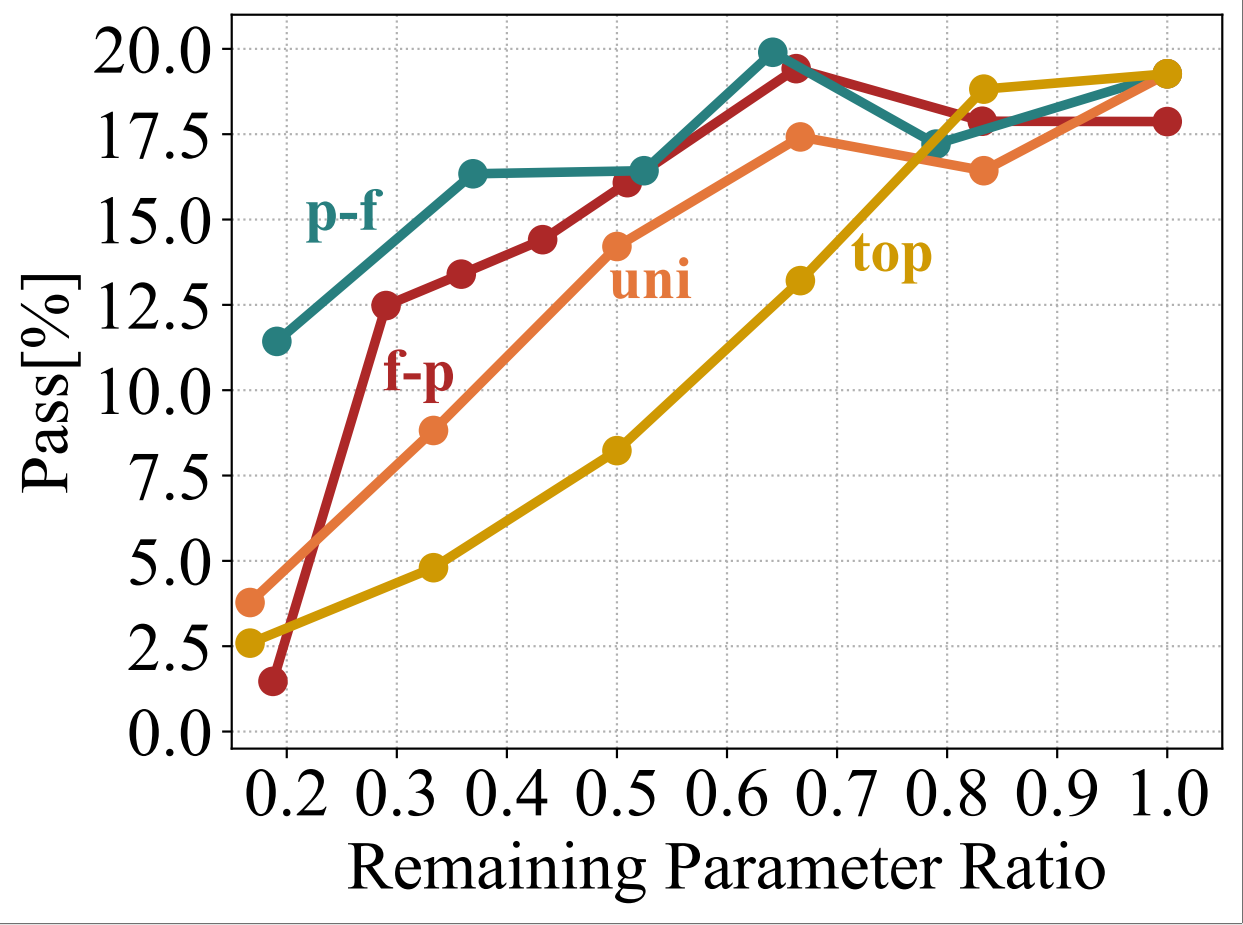}
		\caption{Pass@100}
	\end{subfigure}
    \caption{Performance after Pruning on the HumanEval Dataset.
    The \textit{f-p} is \lm-4B-K2 finetuned with load concentration loss and then pruned with expert frequency.
    The \textit{p-f} is \lm-4B-K2 that is pruned with expert frequency first and then finetuned on python corpus.
    The \textit{uni} and \textit{top} is \lm-4B-K2 that is pruned by uniformly dropping layers or dropping the top layers.
    }
    \label{fig:ratio-pass}
\end{figure}

\begin{figure}[]
    \centering
	\begin{subfigure}{0.3\textwidth}
		\centering
		\includegraphics[width=\textwidth]{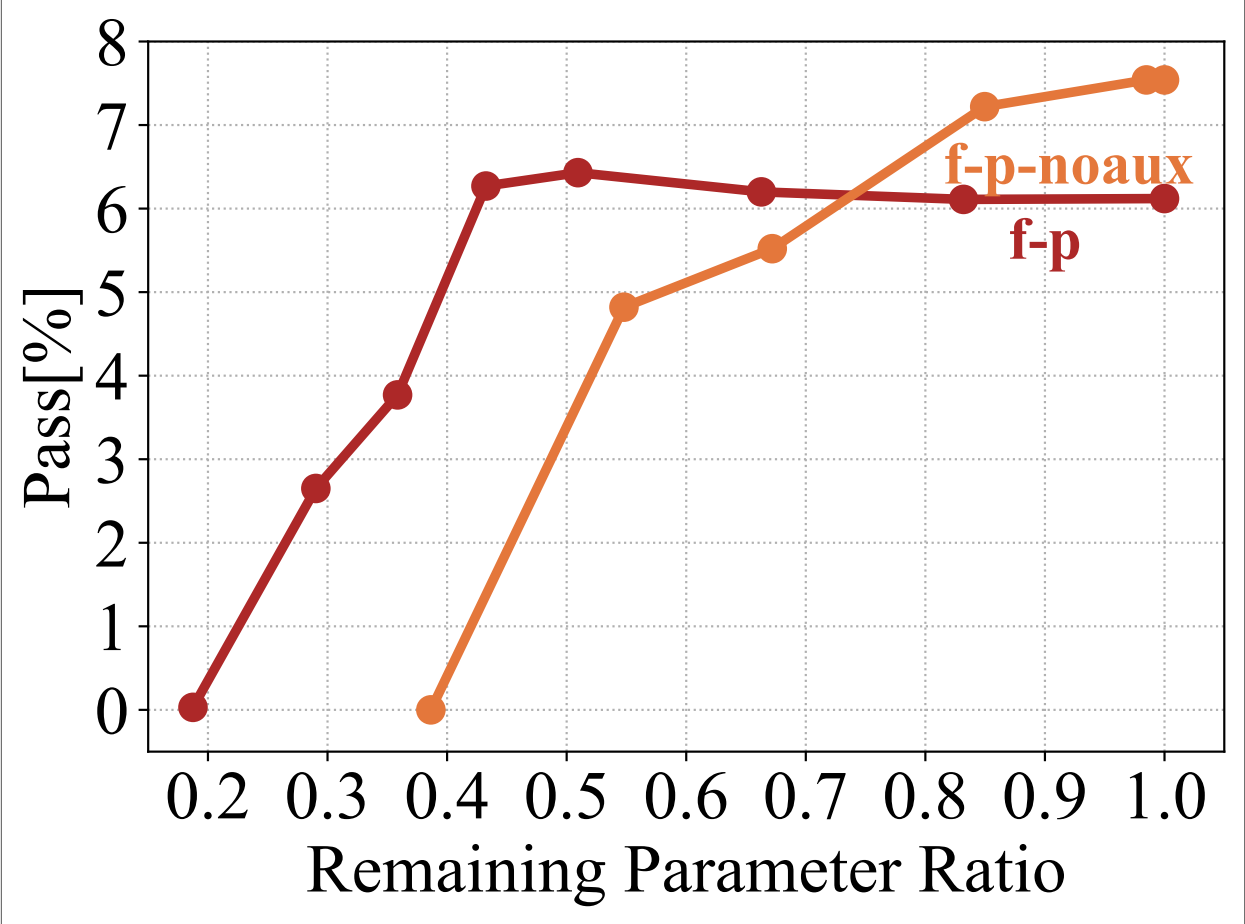}
		\caption{Pass@1}
	\end{subfigure}
	\quad
	\begin{subfigure}{0.3\textwidth}
		\centering
		\includegraphics[width=\textwidth]{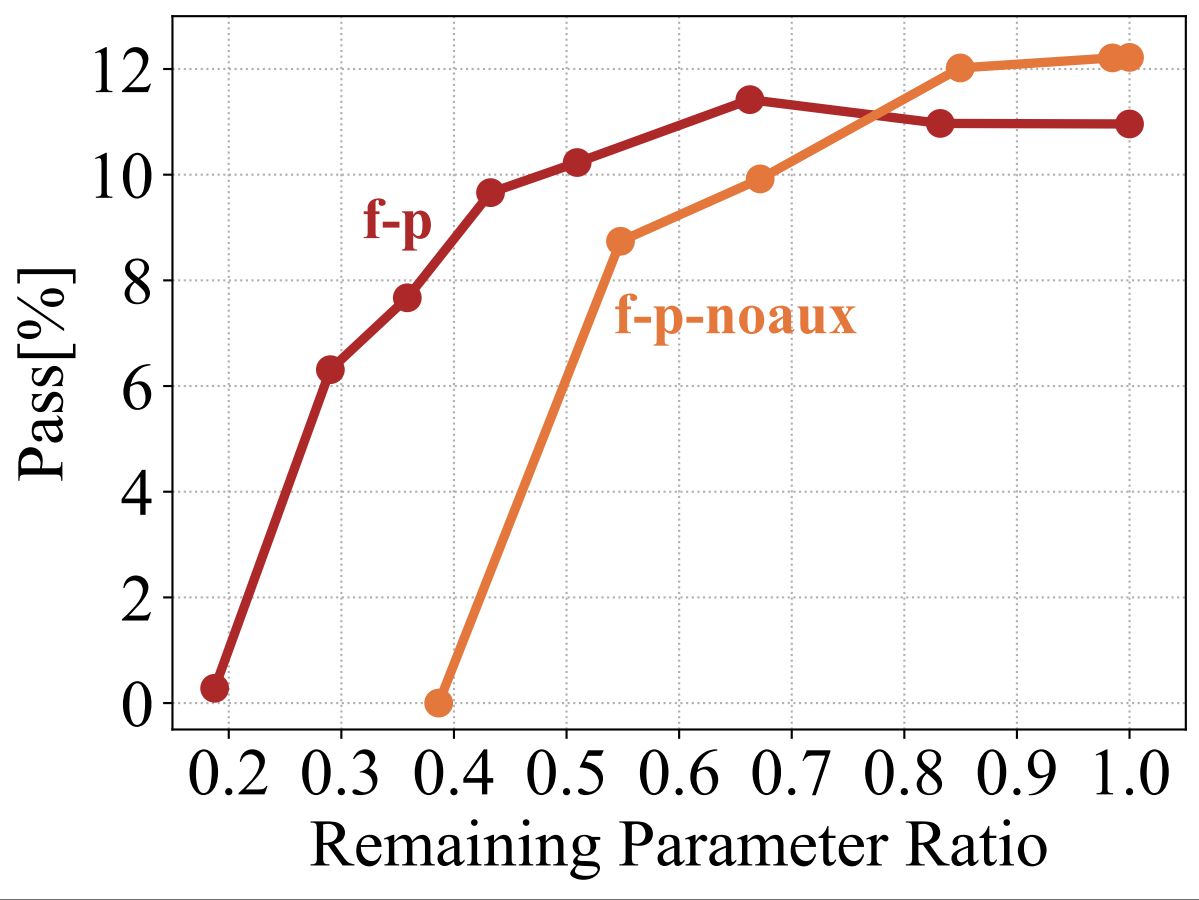}
		\caption{Pass@10}
	\end{subfigure}
	\quad
	\begin{subfigure}{0.3\textwidth}
		\centering
		\includegraphics[width=\textwidth]{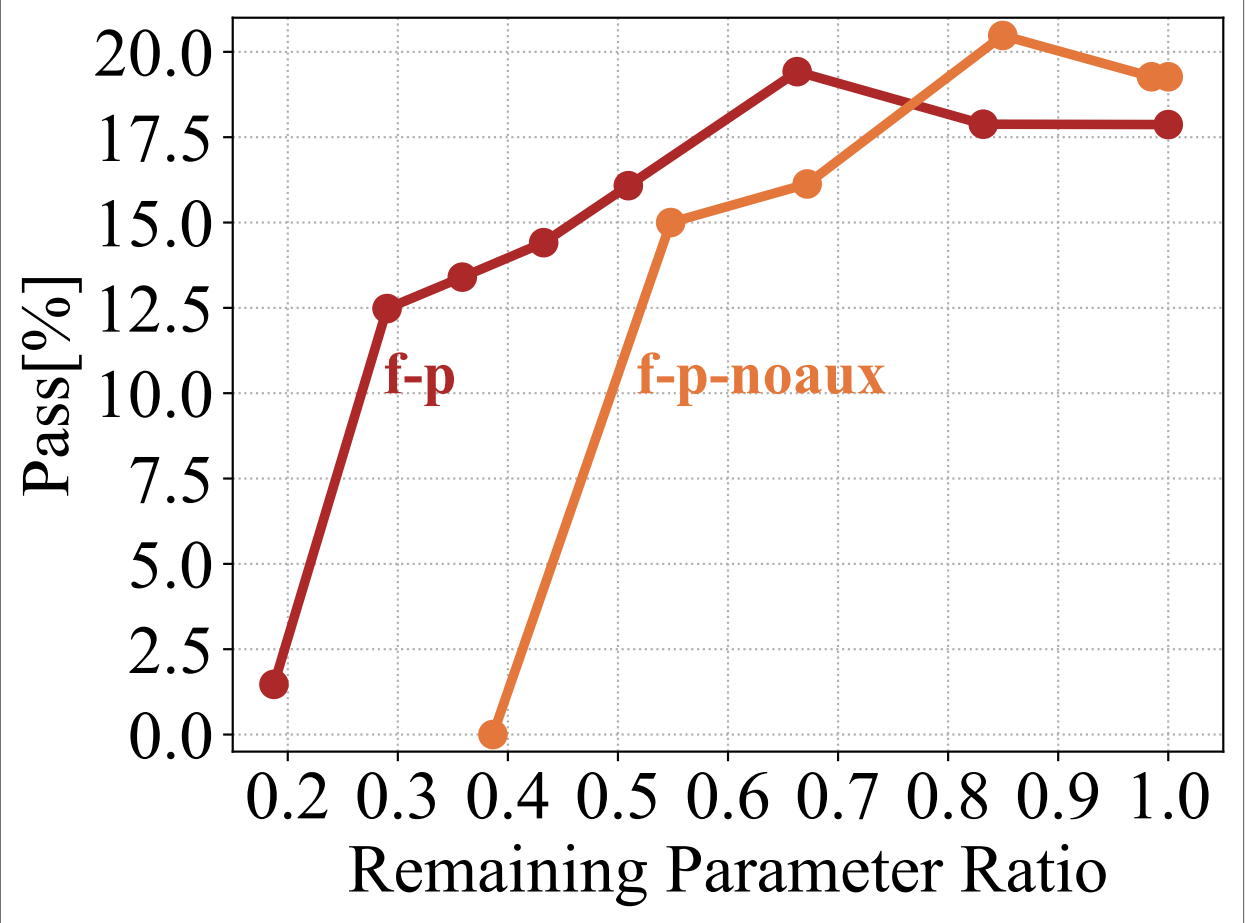}
		\caption{Pass@100}
	\end{subfigure}
    \caption{
    Ablation study for the load concentration loss.
    The \textit{f-p\_noaux} is \lm-4B-K2 finetuned without load concentration loss and then pruned with expert frequency.
    }
    \label{fig:load_concentration_ablation}
\end{figure}

\paragraph{Finetuning}
We finetune our pretrained model on a 15B token subset of the GitHub-code-clean dataset~\citep{tunstall2022natural} containing only Python code.
When we finetune the original model, we add the load concentration loss introduced in Section~\ref{sec:load_concentration} with a weight of $0.001$.
More details for finetuning can be found in Appendix~\ref{app:finetuning}.

\paragraph{Pruning}

We count the activation frequency of each expert on the evaluation set to get the correlation between experts and the targeted domain. 
We then normalize the frequency by dividing the maximum frequency inside each layer. 
After that, we set a threshold $\tau$ and pruned all the modules whose normalized frequency fell below the threshold.

\paragraph{Baselines}
We compare our pruning method with standard layer-wise pruning methods~\citep{sajjad2023effect}.
Specifically, we compared dropping top and uniformly dropping layers with different ratios.

\paragraph{Evaluation}
We test our Pruned MoLM-4B-K2 Model on the HumanEval dataset \cite{chen2021evaluating}.
Figure \ref{fig:ratio-pass} illustrates the correlation between pass@k metrics and the remaining parameter ratios.
We observe that pruning unnecessary modules does not significantly impact the results. 
We can prune 40\% to 50\% of the parameters without sacrificing performance, while the layer-wise pruning method results in a visible performance drop and consistently under-performance module pruning;
2) Our fine-tuning then pruning schema achieves similar performance as pruning then fine-tuning method, while only needs to fine-tune the model once;
3) After fine-tuning with load concentration loss, there are significant disparities in the distribution of modules, with around half of the modules' activation frequency being less than  0.3\% of the most frequently used experts.

Figure \ref{fig:load_concentration_ablation} shows the ablation of load concentration loss. 
Without the load concentration loss, the model performs better for a high remaining parameter ratio (>75\%).  
However, the loss of load concentration enables a better performance from 30\% to 75\% remaining parameter ratio.
These results show that our load concentration loss could enable a more flexible deployment strategy for devices with different capacities.
More finetuning results can be found in Appendix~\ref{app:finetuning}.

\section{Conclusion and Limitations}
\paragraph{Conclusion} 
In this paper, we propose a new modular architecture, \model, and its associated methods for module manipulation. 
\model includes several new components: a new stick-breaking attention heads, a new mutual information load balancing loss for pretraining, and a new load concentration loss for finetuning.
Based on \model, we pretrained a new language model, \lm.
Our experiment result shows the promising abilities of \lm: 
1) It achieves the same performance as dense LLMs with lower latency (50\%) and a smaller memory footprint; thus, it improves the throughput to more than 2 times;
2) It is less susceptible to catastrophic forgetting after finetuning the entire model on a new domain, and it could also be easily extended with new modules to learn a new language;
3) It can be finetuned on a downstream task to specialize a subset of modules on the task and the unused modules can be pruned without sacrificing the performance.

\paragraph{Limitations} Although it is more efficient, the \lm uses more parameters to achieve performance comparable to dense models. 
We believe this is due to the optimization difficulty caused by the discrete gating decision in SMoE.
How optimal gating can be learned remains an open question.









\bibliographystyle{plainnat}
\bibliography{reference}


\appendix


\section{Stick-breaking Self-attention} \label{sec:sbproof}
To restate Equation \eqref{eq:stick-breaking},
\begin{equation}
    p_{i,t} = \beta_{i,t} \prod_{i < j \leq t} \left( 1 - \beta_{j,t} \right) 
\end{equation} 
We can view $p_{t}$ as a distribution over $i$.
The cumulative mass function for attending from timestep $t$ to $k$ timesteps in the past is then,
\begin{align}
    &\sum_{t - k \leq i \leq t} \left[ \beta_{i,t} \prod_{i < j \leq t} \left( 1 - \beta_{j,t} \right) \right] \label{eq:sbcmf}\\
    &= \beta_{t,t} + \beta_{t-1,t}(1 - \beta_{t, t})  + \cdots +  \beta_{t-k,t} \prod_{t - k < j \leq t} \left( 1 - \beta_{j,t} \right)
\intertext{If we let $\alpha_{i,j} = 1 - \beta_{i,j}$ for all $i,j$}
&= (1 - \alpha_{t,t}) + (1 - \alpha_{t-1,t})~\alpha_{t, t}  + \cdots +  (1 - \alpha_{t-k,t}) \prod_{t - k < j \leq t} \alpha_{j,t}
\intertext{and by telescoping cancellation,}
& = 1 - \prod_{t - k \leq j \leq t} \alpha_{j,t}\\
& = 1 -  \prod_{t - k \leq j \leq t} (1  - \beta_{j,t})
\end{align}
Since $\beta_{j,t} \in (0, 1)$, $(1 - \beta_{j,t}) \in (0, 1)$.
Therefore, as $k$ increases (longer context window), Equation \eqref{eq:sbcmf} approaches 1.

\section{Pretraining Hyperparameters}
\label{app:pretrain}
\begin{table}[h]
    \centering
    \caption{Pretraining Hyperparameters}
    \begin{tabular}{c|c}
        \toprule
        Optimizer & AdamW \\
        Maximum Learning Rate & 3e-4 \\
        Minimum Learning Rate & 3e-5 \\
        Weight Decay & 0.01 \\
        Gradient Clipping & 1.0 \\
        Training Tokens & 360B \\
        Number of Epochs & 1 \\
        Warmup Tokens & 3B \\
        Batch Tokens & 3M \\
        Input Sequence Length & 512 \\
        \bottomrule
    \end{tabular}
\end{table}

\section{Continual Learning Experiment Details}
\label{app: continual details}


\subsection{Dataset Details}

CC-100~\citep{wenzek-etal-2020-ccnet, conneau-etal-2020-unsupervised} is a multilingual corpus that collects texts from CommonCrawl in 100 languages. For joint training setting, we incorporate newly learned languages along with English texts at a ratio of 4:1. Additionally, in order to demonstrate our efficiency, we limit our training to only 6 billion tokens.

\subsection{Benchmark Details}

The benchmark, mLAMA~\citep{kassner2021multilingual} offers fill-in-the-blank questions that require comprehensive knowledge, such as birthplace, capital city, and so on. As previous works, we randomly replaced the blank with two incorrect answers among all candidates in mLAMA to test whether the correct sentence is the most unperplexed one.

\subsection{Baselines Details}

\textbf{GPT-Neo}~\citep{gao2020pile} is a GPT-2 like large language dense model, pretrained on the pile~\citep{gao2020pile}, the same corpus we used.

\textbf{LoRA}~\citep{hu2021lora} is a state-of-the-art method for parameter-efficient large language models adaptation. 
In practice, the rank in the LoRA method is frequently referred to using small numbers, such as $16$ or $32$. 
In our experiment, we employ LoRA in both a small and medium rank setting to regulate the number of trainable parameters. 

\subsection{Continual Training Hyperparameters}


\begin{table}[H]
  \centering
  \small
  \caption{Continual Training Hyperparameters}
  \begin{tabular}{l|l}
    \toprule
    Training Tokens & 6B \\
    Number of Epochs & 1 \\
    Warmup Tokens & 3B \\
    Batch Tokens & 327680 \\
    Learning Rate & 3e-4 \\
    Learning Rate Scheduler & Constant \\
    \bottomrule 
  \end{tabular}
\end{table}

\subsection{Continual Training Ablation} \label{app: ablation: continual}

Table~\ref{tab: zero en ab} presents the findings from the continual training ablation study. The results indicate that: 1) With proper routing regularization, a partially trainable router strictly outperforms a fully trainable one. 2) Generally, joint training tends to yield superior overall performance compared to lifelong training, and exceptions only occur in new language abilities with low regularization.

\begin{table}[h]
  \small
  \caption{Comparison of different continual learning methods}
  \label{tab: zero en ab}
  \centering
  \small
  \begin{tabular}{lllc|cc}
    \toprule
    & Trainable & Router & \multicolumn{2}{c}{Continual Training (vi)} \\
    & Params & Tuning & en & vi \\
    \midrule
    \multirow{5}{*}{Continual Joint Pre-Training}  & 164M & Full + load balancing & 79.5(-1.4) & 64.5(+6.1)  \\
     & 151M & Partial & 79.5(-1.4) & \textbf{65.1}(+6.7)  \\
     & 151M & Partial + $0.25$ Rout Reg & 80.1(-0.8) & 65.0(+6.6) \\
     & 151M & Partial + $0.50$ Rout Reg & 80.3(-0.6) & 65.0(+6.6) \\
     & 151M & Partial + $1.00$ Rout Reg & \textbf{80.5}(-0.4) & 64.1(+5.7) \\
     \midrule
    \multirow{5}{*}{Continual Lifelong Pre-Training} & 164M & Full + load balancing & 73.8(-7.1) & 66.0(+7.6) \\
    & 151M & Partial & 74.5(-6.4) & \textbf{68.0}(+9.6) \\
     & 151M & Partial + $0.25$ Rout Reg & 76.0(-4.9) & 64.8(+6.4) \\
     & 151M & Partial + $0.50$ Rout Reg & 76.3(-4.6) & 64.5(+6.1) \\
     & 151M & Partial + $1.00$ Rout Reg & \textbf{78.7}(-2.2) & 63.0(+4.6) \\
    \bottomrule 
  \end{tabular}
  \vspace{-0.2in}
\end{table}

\section{Finetuning and Pruning} \label{app:finetuning}

\subsection{Finetuning Hyperparameters}
\begin{table}[H]
  \small
  \centering
  \caption{Finetuning Hyperparameters}
  \begin{tabular}{l|l}
    \toprule
    Finetuning tokens & 15B \\
    Number of Epochs & 2 \\
    Warmup Tokens & 3B \\
    Batch tokens & 1.5M \\
    Learning Rate & 5e-5 \\
    Learning Rate Scheduler & Constant \\
    \bottomrule 
  \end{tabular}
\end{table}

\subsection{Pruning Ablation} \label{app: ablation: pruning}


\begin{figure}[h]
    \centering
	\begin{subfigure}{0.3\textwidth}
		\centering
		\includegraphics[width=\textwidth]{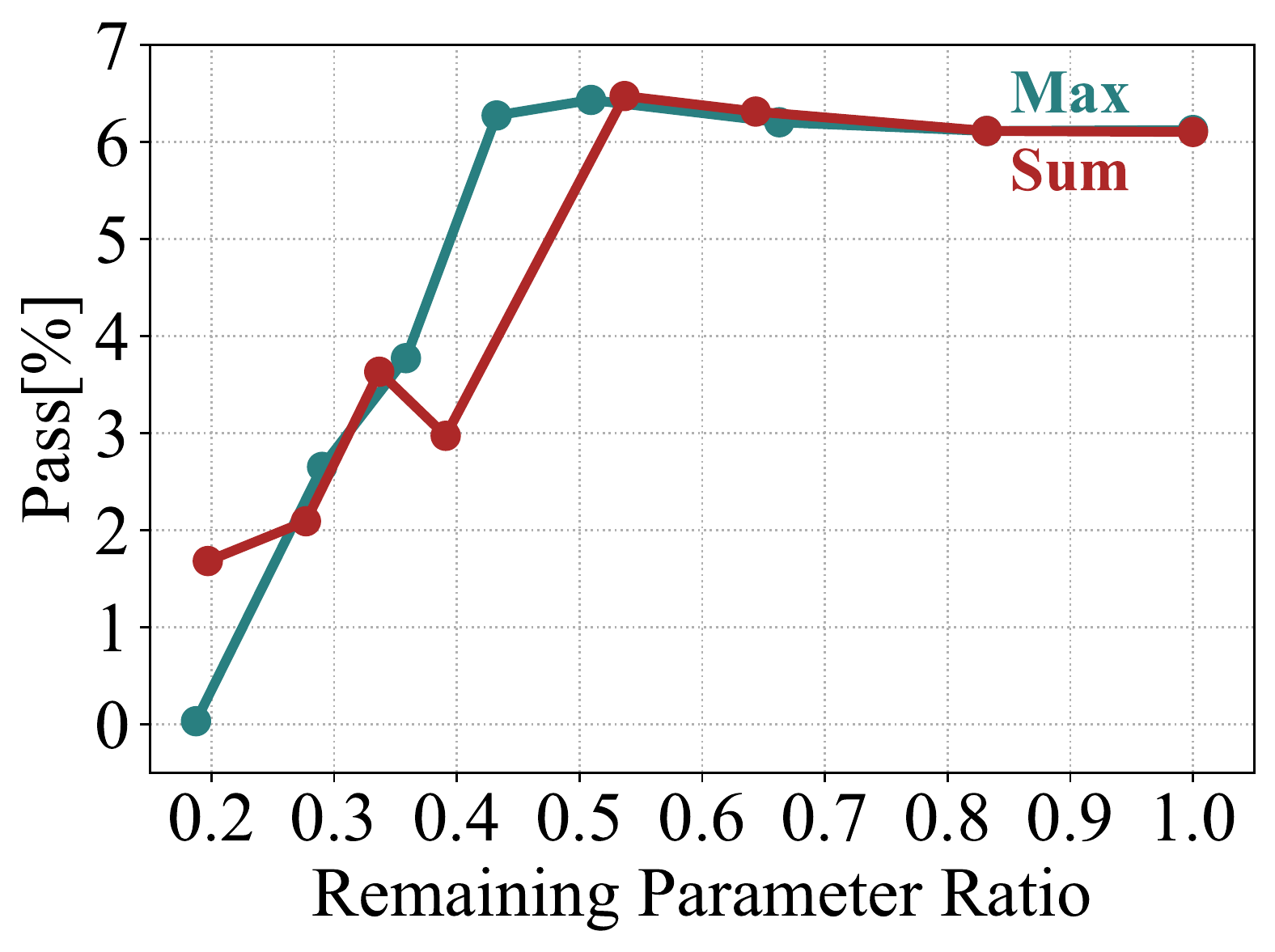}
		\caption{ratios-pass@1}
	\end{subfigure}
	\quad
	\begin{subfigure}{0.3\textwidth}
		\centering
		\includegraphics[width=\textwidth]{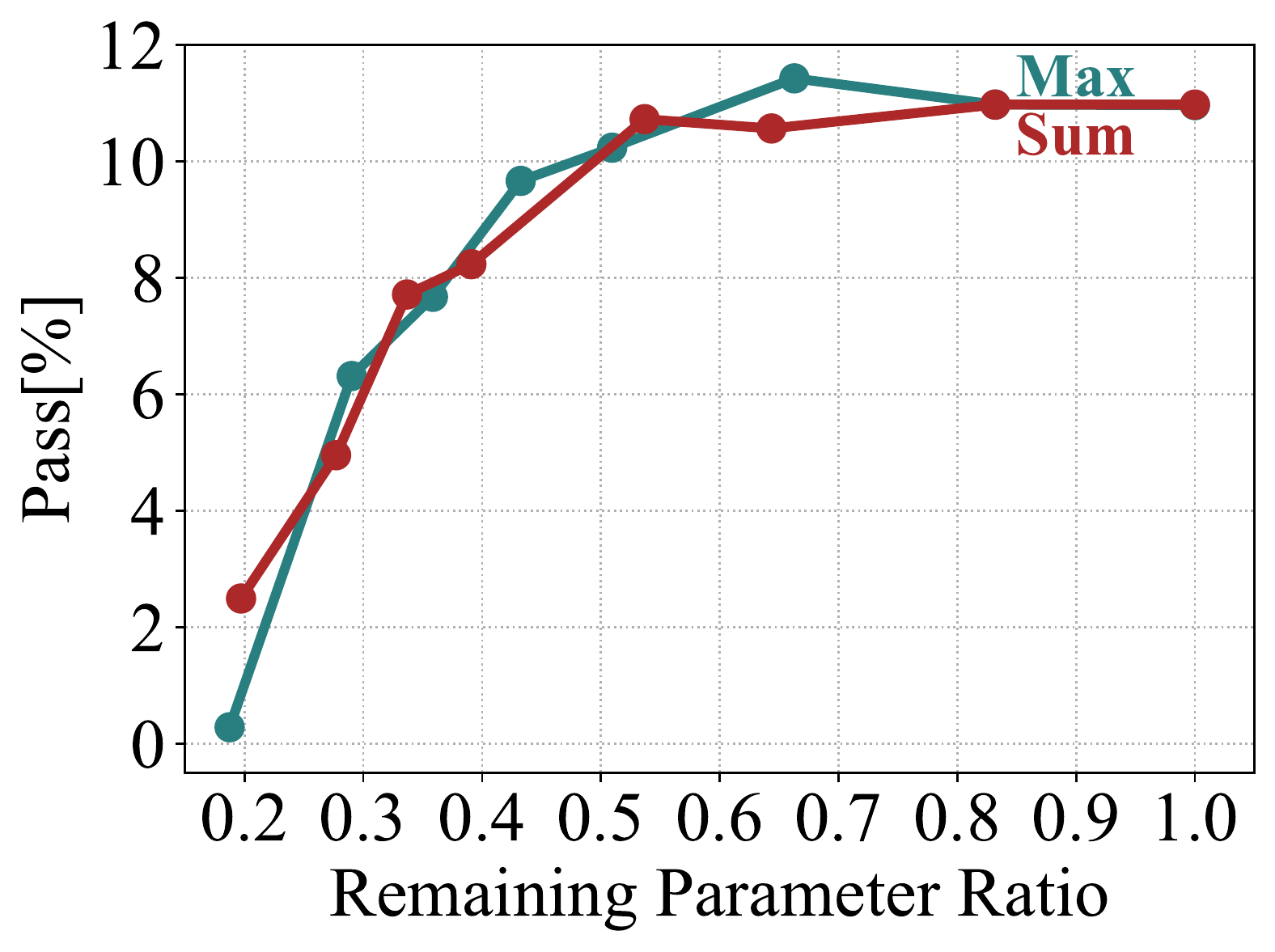}
		\caption{ratios-pass@10}
	\end{subfigure}
	\quad
	\begin{subfigure}{0.3\textwidth}
		\centering
		\includegraphics[width=\textwidth]{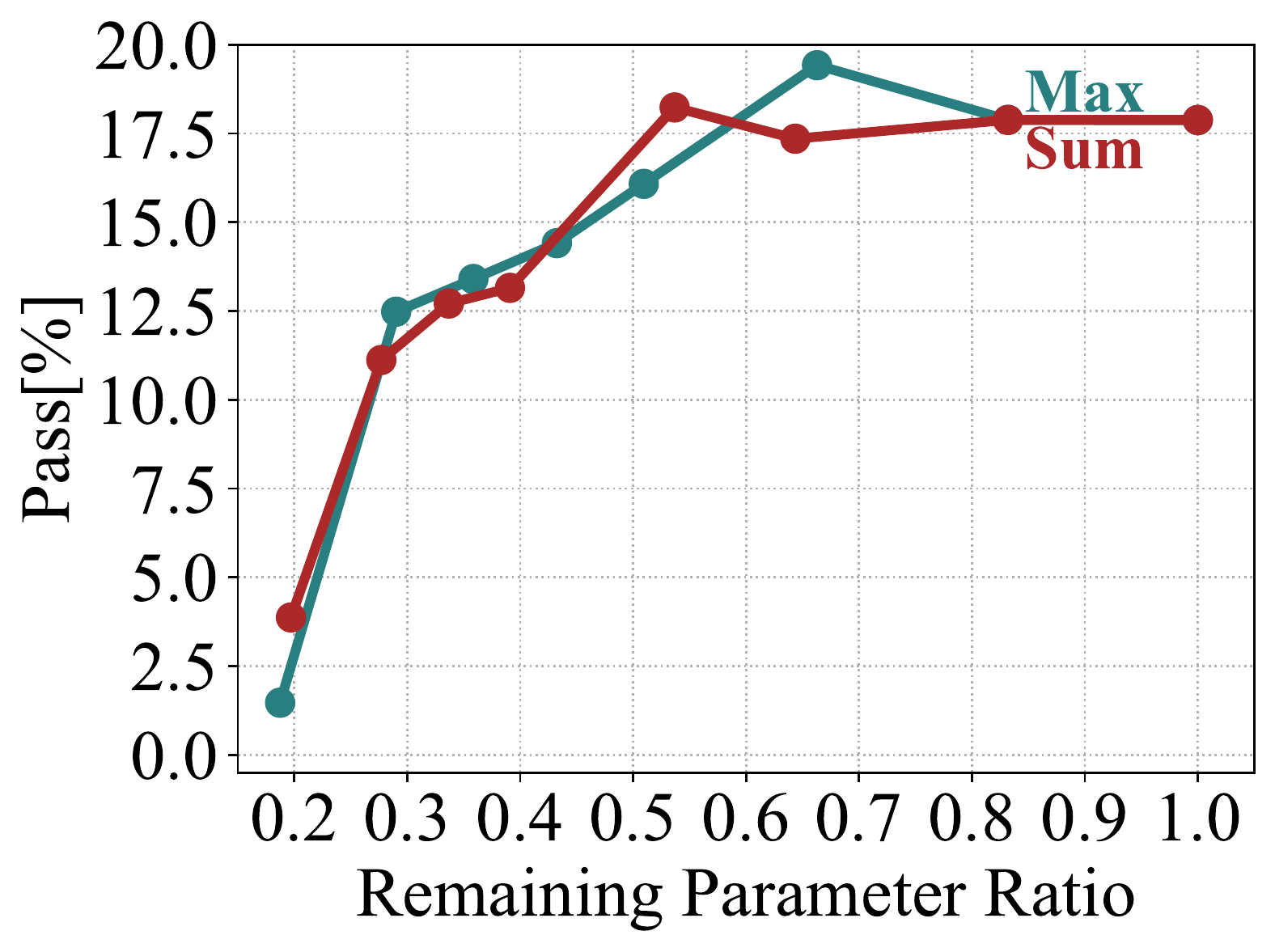}
		\caption{ratios-pass@100}
	\end{subfigure}
    \caption{
    Performances of different pruning methods. 
    \textit{Max} means that the expert frequency is normalized with the maximum frequency inside each layer.
    \textit{Sum} means that the expert frequency is normalized with the total activation frequency of each layer.
    }
    \label{fig:different_pruning}
\end{figure}


Fig.~\ref{fig:different_pruning} compares the different pruning methods. 
While both methods leverage the activation frequency of each expert to prune the model, how to normalize the frequency still makes some differences.
\textit{Max} means that the expert frequency is normalized with the maximum frequency inside each layer.
\textit{Sum} means that the expert frequency is normalized with the total activation frequency of each layer. 
We found that the two methods achieve comparable results for pass@10 and pass@100, but the \textit{Max} achieves better results for pass@1.
Thus we choose the \textit{Max} method as the major results reported in the paper.

\end{document}